\documentclass[11pt]{article}

\usepackage[margin=1in]{geometry}
\usepackage{booktabs}
\usepackage{graphicx}
\usepackage[hidelinks]{hyperref}
\usepackage{microtype}
\usepackage{amsmath}
\usepackage{enumitem}
\usepackage[table]{xcolor}
\usepackage{float}
\usepackage{tabularx}

\title{When JSON Is Not Enough: Semantic Reliability of Schema-Constrained LLM Ordering Agents}
\author{Yin Li\\
University of Birmingham\\
\texttt{kxl474@student.bham.ac.uk}}
\date{May 2026}

\newcommand{\orderbench}{\textsc{OrderBench}}
\begin{document}
\maketitle

\begin{abstract}
LLM agents are increasingly used as transaction compilers: a user states an
intent in natural language, and the model emits a structured object that an API
can execute. JSON Schema and provider-level structured-output modes are useful
because they remove a large class of parse failures, but they do not by
themselves decide whether the object is a safe, faithful transaction. We
introduce \orderbench, a deterministic benchmark for restaurant ordering agents
that separates syntactic validity, schema validity, status decisions, exact item
semantics, constraint preservation, and unsafe acceptances. Across 2,400
Nebius Token Factory calls to four open models in prompt-only and JSON-schema
modes, we find that schema-valid output can still have large semantic error
rates. In the strongest model, both modes achieve 100\% schema validity, yet
semantic success remains near 80\%; in weaker models, schema-valid unsafe
acceptances occur in double digits. The result is a concrete engineering
warning: structured output is a necessary interface layer, not a substitute for
domain verification and fail-closed execution.
\end{abstract}

\section{Introduction}
Production LLM agents often sit between unconstrained natural language and
strict transactional APIs. In food ordering, retail support, travel booking,
health intake, and finance operations, the model is not merely formatting text:
it is compiling a user's request into an object that may reserve inventory,
charge money, or trigger a safety-relevant workflow. The practical failure mode
is therefore not only malformed JSON. It is well-formed JSON that says the wrong
thing.

Provider structured-output modes and constrained decoding have made real
progress on the formatting problem. JSON Schema Draft 2020-12 defines a standard
contract language for JSON data~\cite{jsonschema2020}, OpenAPI 3.1 aligns API
schemas with JSON Schema~\cite{openapi31}, and OpenAI's Structured Outputs
announcement reports perfect adherence on its internal complex-schema
evaluations for a supported model~\cite{openai2024structured}. Systems such as
XGrammar further optimize grammar-constrained generation for structured
outputs~\cite{dong2024xgrammar}. These advances are valuable, but their success
criterion is usually structural: can the output be parsed, and does it match the
declared type-level shape?

This paper studies the next layer: semantic executability under a domain
contract. We focus on restaurant ordering because the domain is compact enough
to evaluate exactly, but rich enough to expose production-relevant edge cases:
modifier negation, modifier scope over multiple items, unavailable catalog
entries, allergen conflicts, dietary constraints, and cases where an allergen is
safe only if a modifier is removed. These are common business rules, not toy
reasoning puzzles.

Our contributions are:
\begin{itemize}[leftmargin=*]
  \item \orderbench, a deterministic 300-case benchmark with a small menu,
  hand-coded oracles, and ten transactional edge-case categories.
  \item A verifier that reports JSON validity, schema validity, status
  correctness, exact item semantics, constraint preservation, semantic success,
  and unsafe acceptance.
  \item A paired evaluation of prompt-only JSON generation versus provider
  JSON-schema mode using the same cases, models, prompts, and temperature.
  \item An empirical demonstration that schema validity can be perfect while
  semantic reliability remains insufficient for direct execution.
\end{itemize}

\section{Related Work}
\paragraph{Tool and function calling.}
Gorilla and APIBench study whether models can select APIs and produce accurate
arguments, emphasizing hallucinated or incorrect calls as a barrier to reliable
tool use~\cite{patil2023gorilla}. The Berkeley Function Calling Leaderboard
systematizes function-calling evaluation across languages, APIs, SQL, relevance
detection, and multi-turn settings~\cite{patil2025bfcl}. \(\tau\)-bench moves
closer to deployment by testing domain-specific tool agents interacting with
users and policies, and reports that even strong function-calling agents remain
inconsistent in realistic domains~\cite{yao2024taubench}. \orderbench{} is
narrower than these benchmarks, but more surgical: it isolates the single-turn
contract compilation step and scores the difference between structurally valid
and domain-correct arguments.

\paragraph{Structured generation.}
JSON Schema is now a common target for structured LLM output. Recent work
benchmarks constrained decoding frameworks against JSON Schema test suites and
structured-output tasks~\cite{geng2025jsonschemabench}, while XGrammar focuses on
efficient grammar execution for structured generation~\cite{dong2024xgrammar}.
Our work assumes that structured generation is useful and asks what remains
unsolved after the structure is valid.

\section{Benchmark}
\orderbench{} defines a menu with seven SKUs, item sizes, default modifiers,
allowed additions/removals, allergens, and dietary tags. Each example contains a
customer utterance and a deterministic oracle object with five top-level fields:
\texttt{status}, \texttt{items}, \texttt{constraints},
\texttt{clarification\_question}, and \texttt{reasons}. The status can be
\texttt{accepted}, \texttt{needs\_clarification}, or
\texttt{rejected\_safety}. The schema forbids additional properties and bounds
quantities to executable integer values.

The 300 examples are evenly distributed across ten categories:
simple exact orders, quantity and size, negated modifiers, scoped modifiers,
shared modifiers, allergen-safe acceptances, allergen conflicts, dietary
conflicts, unavailable items, and unavailable modifiers. Utterances are
deterministic paraphrases of hand-written transactional templates; no model is
used to create labels. This design trades breadth for auditability: every oracle
can be inspected and regenerated from source.

\section{Evaluation}
For each model and mode, we run all 300 cases at temperature zero through the
Nebius Token Factory OpenAI-compatible API~\cite{nebius2026tokenfactory}. The
two modes are:
\begin{itemize}[leftmargin=*]
  \item \textbf{Prompt-only}: the prompt asks the model to return only JSON and
  includes the schema as text.
  \item \textbf{JSON schema}: the same prompt plus provider
  strict \texttt{json\_schema} response formatting.
\end{itemize}

We evaluate four models available from the provider at run time: GPT-OSS
120B-fast, Qwen3-30B-A3B, Llama-3.1-8B, and Gemma-2-2B. The exact provider
model identifiers are recorded in the released JSONL outputs. A compatibility
pass excluded a fifth candidate that frequently exhausted the response budget
and failed parsing under this prompt. Raw model responses, provider usage,
latency, parsed objects, and verifier results are written as JSONL.

The main metric is \textbf{semantic success}. For accepted orders, it requires
correct status, exact multiset equality over SKU, quantity, size, additions,
removals, and special instructions, plus exact preservation of stated allergens
and dietary constraints. For non-accepted orders, it requires correct status,
empty executable items, and exact constraints. This is intentionally strict: it
measures whether the model's object can be used as a drop-in transaction
contract, not whether the answer is approximately useful. Because exact
equality does not imply equal severity, we also report a multi-label error
taxonomy: safety failures, catalog hallucinations, quantity/size errors, scope
split errors, negation errors, and allergen-preservation errors.

The most important risk metric is \textbf{unsafe acceptance}. We count an unsafe
acceptance only when the model emits \texttt{status=accepted} for an object that
the verifier says must not be sent to execution: an allergen conflict, a dietary
conflict, an unavailable item, an unavailable modifier, or an accepted item that
still conflicts with a stated allergen. Ordinary exactness misses, such as
omitting a default size, are semantic failures but are not counted as unsafe.
For paired comparisons between prompt-only and JSON-schema modes, we report the
schema-minus-prompt rate difference, a paired bootstrap 95\% confidence
interval over cases, and an exact McNemar test on discordant paired outcomes.

\section{Results}
Table~\ref{tab:main} reports the full paired evaluation. The key pattern is that
JSON and schema validity are much easier than semantic reliability. Structured
mode often improves schema validity for smaller models, but it does not
guarantee better business decisions or safer executable objects.

The strongest evaluated model, GPT-OSS 120B-fast, reaches 100\% schema validity
in both modes, but semantic success is 83.0\% in prompt-only mode and 81.3\% in
JSON-schema mode. Qwen3-30B-A3B also reaches 100\% schema validity in both
modes, while semantic success is only 31.3\% and 30.7\%, respectively, and
unsafe acceptance remains around 15\%. For Llama-3.1-8B, structured output
raises schema validity from 68.7\% to 100.0\% and reduces unsafe acceptance from
13.3\% to 8.3\%, but semantic success remains 36.0\%. Gemma-2-2B is the most
severe counterexample: JSON-schema mode produces 100\% schema-valid objects, yet
semantic success is 2.0\% and unsafe acceptance is 41.7\%.

\begin{table}[H]
\centering
\small
\begin{tabular}{llrrrrrr}
\toprule
Model & Mode & $n$ & JSON & Schema & Status & Semantic & Unsafe \\
\midrule
Qwen3-30B-A3B & json\_schema & 300 & 100.0 & 100.0 & 78.0 & 30.7 & 14.3 \\
Qwen3-30B-A3B & prompt\_only & 300 & 100.0 & 100.0 & 78.7 & 31.3 & 15.0 \\
Gemma-2-2B & json\_schema & 300 & 100.0 & 100.0 & 61.0 & 2.0 & 41.7 \\
Gemma-2-2B & prompt\_only & 300 & 99.7 & 91.7 & 60.0 & 0.0 & 45.0 \\
Llama-3.1-8B & json\_schema & 300 & 100.0 & 100.0 & 69.7 & 36.0 & 8.3 \\
Llama-3.1-8B & prompt\_only & 300 & 100.0 & 68.7 & 74.3 & 31.7 & 13.3 \\
gpt-oss-120b-fast & json\_schema & 300 & 100.0 & 100.0 & 98.7 & 81.3 & 0.0 \\
gpt-oss-120b-fast & prompt\_only & 300 & 100.0 & 100.0 & 98.0 & 83.0 & 0.3 \\
\bottomrule
\end{tabular}

\caption{Main results. Values other than \(n\) are percentages. ``Semantic''
is exact domain-contract success; ``Unsafe'' is accepting an order that should
not be executable.}
\label{tab:main}
\end{table}

Table~\ref{tab:category} aggregates semantic success by category and mode. The
hard categories are not generic JSON formatting tasks. They are domain boundary
tasks: item availability, modifier availability, and user constraints that must
change the execution status. Because this table averages over models, Appendix
Table~\ref{tab:heatmap} provides the model-by-category heatmap used to check
whether a category-level effect is broad or driven by one weak model.

\begin{table}[H]
\centering
\small
\begin{tabular}{llrrr}
\toprule
Category & Mode & $n$ & Semantic & Unsafe \\
\midrule
allergen-safe & json\_schema & 120 & 30.8 & 6.7 \\
allergen-safe & prompt\_only & 120 & 30.8 & 13.3 \\
allergen-conflict & json\_schema & 120 & 35.0 & 35.8 \\
allergen-conflict & prompt\_only & 120 & 36.7 & 36.7 \\
dietary-conflict & json\_schema & 120 & 28.3 & 25.8 \\
dietary-conflict & prompt\_only & 120 & 34.2 & 27.5 \\
negation & json\_schema & 120 & 33.3 & 0.0 \\
negation & prompt\_only & 120 & 26.7 & 0.0 \\
qty/size & json\_schema & 120 & 49.2 & 0.0 \\
qty/size & prompt\_only & 120 & 48.3 & 0.0 \\
scope & json\_schema & 120 & 19.2 & 0.0 \\
scope & prompt\_only & 120 & 25.0 & 0.0 \\
shared & json\_schema & 120 & 39.2 & 0.0 \\
shared & prompt\_only & 120 & 34.2 & 0.0 \\
simple & json\_schema & 120 & 43.3 & 0.0 \\
simple & prompt\_only & 120 & 45.8 & 0.0 \\
bad-item & json\_schema & 120 & 48.3 & 51.7 \\
bad-item & prompt\_only & 120 & 40.8 & 58.3 \\
bad-modifier & json\_schema & 120 & 48.3 & 40.8 \\
bad-modifier & prompt\_only & 120 & 42.5 & 48.3 \\
\bottomrule
\end{tabular}

\caption{Category-level results aggregated across models. Values are
percentages.}
\label{tab:category}
\end{table}

Table~\ref{tab:taxonomy} separates exact semantic failures by operational
severity. The labels are multi-label rather than mutually exclusive; for
example, accepting a menu-external modifier is both a catalog error and an
unsafe acceptance. The table makes the strictness of semantic success easier to
interpret: quantity/size errors and scope errors are common, but the deployment
critical categories are unsafe acceptances and catalog hallucinations.

\begin{table}[H]
\centering
\small
\begin{tabular}{lrrrrrrr}
\toprule
Mode & $n$ & Unsafe & Catalog & Qty/size & Scope & Negation & Allergen \\
\midrule
json\_schema & 1200 & 16.1 & 16.2 & 5.4 & 8.1 & 6.7 & 4.7 \\
prompt\_only & 1200 & 18.4 & 17.2 & 6.8 & 7.5 & 7.3 & 4.8 \\
\bottomrule
\end{tabular}

\caption{Multi-label error taxonomy aggregated across models. Values are
percentages of all cases in each mode. Categories do not sum to 100\%.}
\label{tab:taxonomy}
\end{table}

\begin{table}[H]
\centering
\small
\begin{tabularx}{\textwidth}{>{\raggedright\arraybackslash}X>{\raggedright\arraybackslash}p{0.18\textwidth}>{\raggedright\arraybackslash}X>{\raggedright\arraybackslash}p{0.14\textwidth}}
\toprule
User utterance & Oracle & Model output & Error type \\
\midrule
``No dairy for me, add cheese sauce to fries.'' & reject for allergen safety & Qwen3 (schema) accepts fries with cheese sauce. & unsafe allergen \\
``Two vegan bowls, only one with avocado.'' & accepted: split into one avocado bowl and one plain bowl & Qwen3 (schema) emits quantity 2 with avocado. & scope split \\
``Classic burger with pineapple.'' & needs clarification & Qwen3 (schema) accepts pineapple as a burger add-on. & catalog hallucination \\
``I need a vegan option, classic burger with extra cheese.'' & needs clarification & Qwen3 (schema) accepts a classic burger with extra cheese. & unsafe dietary \\
``I'm allergic to peanuts. Get me satay noodles.'' & reject for allergen safety & Gemma (schema) accepts satay noodles. & unsafe allergen \\
\bottomrule
\end{tabularx}

\caption{Representative schema-valid failures. These examples illustrate why
semantic exactness, unsafe acceptance, and catalog errors are reported
separately.}
\label{tab:examples}
\end{table}

Table~\ref{tab:paired} reports the paired uncertainty estimates promised in the
evaluation protocol. The headline semantic differences for GPT-OSS and Qwen are
small and statistically indistinguishable from zero. For Llama-3.1-8B,
JSON-schema mode improves schema validity and lowers unsafe acceptance, but the
semantic-success gain is not significant at the 5\% level. Gemma's
semantic-success improvement is statistically nonzero but remains practically
inadequate at 2.0\% absolute success. Unsafe acceptance shows a different
pattern: JSON-schema mode significantly reduces unsafe acceptance for Gemma and
Llama-3.1-8B, but does not eliminate the risk.

\begin{table}[H]
\centering
\small
\begin{tabular}{llrrrr}
\toprule
Model & Metric & Prompt & Schema & $\Delta$ [95\% CI] & $p$ \\
\midrule
gpt-oss-120b-fast & Semantic & 83.0 & 81.3 & -1.7 [-5.0, +1.7] & 0.458 \\
gpt-oss-120b-fast & Unsafe & 0.3 & 0.0 & -0.3 [-1.0, +0.0] & 1.000 \\
Qwen3-30B-A3B & Semantic & 31.3 & 30.7 & -0.7 [-3.3, +2.0] & 0.804 \\
Qwen3-30B-A3B & Unsafe & 15.0 & 14.3 & -0.7 [-2.0, +0.7] & 0.625 \\
Llama-3.1-8B & Semantic & 31.7 & 36.0 & +4.3 [-0.7, +9.3] & 0.117 \\
Llama-3.1-8B & Unsafe & 13.3 & 8.3 & -5.0 [-8.0, -2.0] & 0.001 \\
Gemma-2-2B & Semantic & 0.0 & 2.0 & +2.0 [+0.7, +3.7] & 0.031 \\
Gemma-2-2B & Unsafe & 45.0 & 41.7 & -3.3 [-6.0, -1.0] & 0.013 \\
\bottomrule
\end{tabular}

\caption{Paired prompt-only versus JSON-schema comparison. \(\Delta\) is
JSON-schema minus prompt-only percentage points. Confidence intervals are paired
bootstrap 95\% intervals over cases; \(p\) is an exact McNemar test. Lower is
better for unsafe acceptance.}
\label{tab:paired}
\end{table}

\section{Discussion}
The experiments support a simple deployment rule: do not let schema-valid model
output directly execute a transaction. A JSON Schema can ensure that
\texttt{items} is an array and \texttt{quantity} is an integer. It cannot ensure
that a requested allergen conflict was rejected, that ``one without onion and one
normal'' was split into two item objects, or that ``pineapple'' was not silently
invented as a modifier. Those checks require a domain verifier.

This does not make structured output unimportant. It makes structured output the
first layer in a larger contract stack. A practical architecture is:
\begin{enumerate}[leftmargin=*]
  \item Require strict structured output to eliminate parse and shape failures.
  \item Validate the object against the business catalog and safety policy.
  \item Fail closed on verifier errors, using clarification rather than
  auto-repair for safety-critical fields.
  \item Log model output and verifier decisions as paired artifacts for
  regression testing.
\end{enumerate}

\section{Limitations}
\orderbench{} is synthetic and single-turn. Its strength is auditability, not
coverage of all restaurant language. The menu is intentionally small, so the
reported rates should not be read as deployment estimates for a particular
vendor or restaurant. We also evaluate one prompt and one provider interface;
prompt engineering, fine-tuning, retrieval over a larger catalog, or a
specialized planner-verifier loop could improve absolute performance. The
central claim is narrower: schema validity alone is not a sufficient reliability
metric for transactional execution.

\section{Conclusion}
Structured-output APIs substantially reduce interface friction, but production
agents need semantic contracts, not only syntactic contracts. \orderbench{}
shows that models can satisfy a strict JSON shape while still emitting unsafe or
incorrect executable orders. For transaction-compiling agents, the engineering
unit of reliability should be the verified domain action.

\appendix
\section{Model-by-Category Heatmap}
Table~\ref{tab:heatmap} breaks the category averages in
Table~\ref{tab:category} down by model and mode. Darker cells indicate higher
semantic success. The heatmap shows, for example, that the low aggregate score
for scope is not solely a Gemma artifact: Qwen and Llama also struggle with
split-item semantics, while GPT-OSS is substantially stronger.

\begin{table}[H]
\centering
\scriptsize
\resizebox{\textwidth}{!}{\begin{tabular}{lrrrrrrrr}
\toprule
Category & gpt-oss-120b-fast P & gpt-oss-120b-fast S & Qwen3-30B-A3B P & Qwen3-30B-A3B S & Llama-3.1-8B P & Llama-3.1-8B S & Gemma-2-2B P & Gemma-2-2B S \\
\midrule
simple & \cellcolor[gray]{0.69}90 & \cellcolor[gray]{0.67}93 & \cellcolor[gray]{0.86}40 & \cellcolor[gray]{0.86}40 & \cellcolor[gray]{0.81}53 & \cellcolor[gray]{0.87}37 & \cellcolor[gray]{1.00}0 & \cellcolor[gray]{0.99}3 \\
qty/size & \cellcolor[gray]{0.65}100 & \cellcolor[gray]{0.65}100 & \cellcolor[gray]{0.85}43 & \cellcolor[gray]{0.82}50 & \cellcolor[gray]{0.82}50 & \cellcolor[gray]{0.84}47 & \cellcolor[gray]{1.00}0 & \cellcolor[gray]{1.00}0 \\
negation & \cellcolor[gray]{0.73}77 & \cellcolor[gray]{0.78}63 & \cellcolor[gray]{0.96}10 & \cellcolor[gray]{0.98}7 & \cellcolor[gray]{0.93}20 & \cellcolor[gray]{0.78}63 & \cellcolor[gray]{1.00}0 & \cellcolor[gray]{1.00}0 \\
scope & \cellcolor[gray]{0.76}70 & \cellcolor[gray]{0.76}70 & \cellcolor[gray]{0.94}17 & \cellcolor[gray]{0.98}7 & \cellcolor[gray]{0.95}13 & \cellcolor[gray]{1.00}0 & \cellcolor[gray]{1.00}0 & \cellcolor[gray]{1.00}0 \\
shared & \cellcolor[gray]{0.70}87 & \cellcolor[gray]{0.71}83 & \cellcolor[gray]{0.91}27 & \cellcolor[gray]{0.91}27 & \cellcolor[gray]{0.92}23 & \cellcolor[gray]{0.90}30 & \cellcolor[gray]{1.00}0 & \cellcolor[gray]{0.94}17 \\
allergen-safe & \cellcolor[gray]{0.76}70 & \cellcolor[gray]{0.74}73 & \cellcolor[gray]{0.81}53 & \cellcolor[gray]{0.82}50 & \cellcolor[gray]{1.00}0 & \cellcolor[gray]{1.00}0 & \cellcolor[gray]{1.00}0 & \cellcolor[gray]{1.00}0 \\
allergen-conflict & \cellcolor[gray]{0.66}97 & \cellcolor[gray]{0.67}93 & \cellcolor[gray]{0.82}50 & \cellcolor[gray]{0.84}47 & \cellcolor[gray]{1.00}0 & \cellcolor[gray]{1.00}0 & \cellcolor[gray]{1.00}0 & \cellcolor[gray]{1.00}0 \\
dietary-conflict & \cellcolor[gray]{0.86}40 & \cellcolor[gray]{0.87}37 & \cellcolor[gray]{1.00}0 & \cellcolor[gray]{1.00}0 & \cellcolor[gray]{0.66}97 & \cellcolor[gray]{0.73}77 & \cellcolor[gray]{1.00}0 & \cellcolor[gray]{1.00}0 \\
bad-item & \cellcolor[gray]{0.65}100 & \cellcolor[gray]{0.65}100 & \cellcolor[gray]{0.90}30 & \cellcolor[gray]{0.86}40 & \cellcolor[gray]{0.88}33 & \cellcolor[gray]{0.81}53 & \cellcolor[gray]{1.00}0 & \cellcolor[gray]{1.00}0 \\
bad-modifier & \cellcolor[gray]{0.65}100 & \cellcolor[gray]{0.65}100 & \cellcolor[gray]{0.85}43 & \cellcolor[gray]{0.86}40 & \cellcolor[gray]{0.91}27 & \cellcolor[gray]{0.81}53 & \cellcolor[gray]{1.00}0 & \cellcolor[gray]{1.00}0 \\
\bottomrule
\end{tabular}
}
\caption{Semantic success by category, model, and mode. P = prompt-only; S =
JSON-schema. Values are percentages; darker cells are higher.}
\label{tab:heatmap}
\end{table}

\bibliographystyle{plain}
\bibliography{references}

@misc{jsonschema2020,
  title = {{JSON Schema Draft 2020-12}},
  author = {{JSON Schema Organization}},
  year = {2022},
  howpublished = {\url{https://json-schema.org/draft/2020-12}},
  note = {Accessed 2026-05-16}
}

@misc{openapi31,
  title = {{OpenAPI Specification v3.1.0}},
  author = {{OpenAPI Initiative}},
  year = {2021},
  howpublished = {\url{https://spec.openapis.org/oas/v3.1.0.html}},
  note = {Accessed 2026-05-16}
}

@misc{openai2024structured,
  title = {Introducing Structured Outputs in the {API}},
  author = {{OpenAI}},
  year = {2024},
  howpublished = {\url{https://openai.com/index/introducing-structured-outputs-in-the-api/}},
  note = {Accessed 2026-05-16}
}

@misc{nebius2026tokenfactory,
  title = {{Nebius Token Factory API Reference}},
  author = {{Nebius}},
  year = {2026},
  howpublished = {\url{https://docs.tokenfactory.nebius.com/api-reference/introduction}},
  note = {Accessed 2026-05-16}
}

@misc{patil2023gorilla,
  title = {Gorilla: Large Language Model Connected with Massive {APIs}},
  author = {Patil, Shishir G. and Zhang, Tianjun and Wang, Xin and Gonzalez, Joseph E.},
  year = {2023},
  eprint = {2305.15334},
  archivePrefix = {arXiv},
  primaryClass = {cs.CL}
}

@misc{yao2024taubench,
  title = {{$\tau$}-bench: A Benchmark for Tool-Agent-User Interaction in Real-World Domains},
  author = {Yao, Shunyu and Shinn, Noah and Razavi, Pedram and Narasimhan, Karthik},
  year = {2024},
  eprint = {2406.12045},
  archivePrefix = {arXiv},
  primaryClass = {cs.AI}
}

@inproceedings{patil2025bfcl,
  title = {The Berkeley Function Calling Leaderboard ({BFCL}): From Tool Use to Agentic Evaluation of Large Language Models},
  author = {Patil, Shishir G. and Mao, Huanzhi and Yan, Fanjia and Ji, Charlie Cheng-Jie and Suresh, Vishnu and Stoica, Ion and Gonzalez, Joseph E.},
  booktitle = {Proceedings of the 42nd International Conference on Machine Learning},
  pages = {48371--48392},
  year = {2025},
  volume = {267},
  series = {Proceedings of Machine Learning Research},
  publisher = {PMLR},
  howpublished = {\url{https://proceedings.mlr.press/v267/patil25a.html}}
}

@misc{dong2024xgrammar,
  title = {{XGrammar}: Flexible and Efficient Structured Generation Engine for Large Language Models},
  author = {Dong, Yixin and Ruan, Charlie F. and Cai, Yaxing and Lai, Ruihang and Xu, Ziyi and Zhao, Yilong and Chen, Tianqi},
  year = {2024},
  eprint = {2411.15100},
  archivePrefix = {arXiv},
  primaryClass = {cs.CL}
}

@misc{geng2025jsonschemabench,
  title = {{JSONSchemaBench}: A Rigorous Benchmark of Structured Outputs for Language Models},
  author = {Geng, Saibo and Cooper, Hudson and Moskal, Micha{\l} and Jenkins, Samuel and Berman, Julian and Ranchin, Nathan and West, Robert and Horvitz, Eric and Nori, Harsha},
  year = {2025},
  eprint = {2501.10868},
  archivePrefix = {arXiv},
  primaryClass = {cs.CL}
}

\end{document}